\documentclass[10 pt, conference, letterpaper]{IEEEtran}  

\usepackage[letterpaper,
            left=54pt,
            right=54pt,
            top=54pt,
            bottom=54pt]{geometry} 

\def\IEEEtitletopspace{18pt}
\usepackage{mathtools}
\usepackage[T1]{fontenc}
\usepackage[font=normalsize]{caption}
\usepackage{xcolor}

\IEEEoverridecommandlockouts                              

\usepackage{graphics} 
\usepackage{epsfig} 
\usepackage{mathptmx} 
\usepackage{times} 
\usepackage{amsmath} 
\usepackage{amssymb}  
\usepackage[T1]{fontenc} 
\usepackage{dblfloatfix}
\usepackage{flushend}
\usepackage{enumitem}
\usepackage{hyperref}
\usepackage{tabularray}
\usepackage{booktabs}
\usepackage{threeparttable}
\usepackage{makecell}

\usepackage{array}  
\usepackage{multirow}  
\usepackage{arydshln}  

\usepackage{textcomp}

\usepackage{eso-pic}

 \AddToShipoutPicture*{\raisebox{27cm}{\noindent \hspace{0.7cm}This work has been submitted to the IEEE for possible publication.}}

 \AddToShipoutPicture*{\raisebox{26.6cm}{\noindent \hspace{-9.77cm}Copyright may be transferred without notice, after which this version may no longer be accessible.}}

\begin{document}

\title{Active Learning Design: Modeling Force Output for Axisymmetric Soft Pneumatic Actuators} 

\author{Gregory M. Campbell, Gentian Muhaxheri, Leonardo Ferreira Guilhoto, Christian D. Santangelo, \\ Paris Perdikaris, James Pikul, and Mark Yim

}

\maketitle

\begin{abstract}

Soft pneumatic actuators (SPA) made from elastomeric materials can provide large strain and large force. The behavior of locally strain-restricted hyperelastic materials under inflation has been investigated thoroughly for shape reconfiguration, but requires further investigation for trajectories involving external force. In this work we model force-pressure-height relationships for a concentrically strain-limited class of soft pneumatic actuators and demonstrate the use of this model to design SPA response for object lifting. We predict relationships under different loadings by solving energy minimization equations and verify this theory by using an automated test rig to collect rich data for n=22 \hbox{Ecoflex 00-30} membranes. We collect this data using an active learning pipeline to efficiently model the design space. We show that this learned material model outperforms the theory-based model and naive curve-fitting approaches. We use our model to optimize membrane design for different lift tasks and compare this performance to other designs. These contributions represent a step towards understanding the natural response for this class of actuator and embodying intelligent lifts in a single-pressure input actuator system.


\end{abstract}


\section{INTRODUCTION}
    
    
    







Soft actuators are promising for physical human-robot interaction in large part due to their compliance. 
Successful control of a soft pneumatic robot requires careful characterization of the soft manipulator, its fluidic elastomer actuators, and the elements that supply fluid energy to predict these reactions \cite{marchese2016dynamics}. Characterization of relevant elements for a built system is laborious, and even intractable for soft actuators with many design parameters. This paper  presents soft pneumatic actuator design characterization for actuation trajectories involving applications with external forces.

Researchers have developed a robust understanding, via sophisticated modeling \cite{melly2021review}, of hyperelastic silicone materials and their reactions to external forces \cite{feng1973contact, feng1975general}. This understanding has allowed others  to characterize the response of inflated silicone membranes to external forces through analytical solutions \cite{yang2021contact} and energy methods \cite{herzig2021model, shi2023modelling}. Energy methods allow general characterization but require solving sets of ordinary differential equations numerically, which is time-consuming and scales poorly during exploration of a parametric design space. For partially restrained (anisotropic) membranes, some have instead relied on simplified load estimation from contact-area assumptions \cite{ambrose2023compact}. Contact-based force transforms are valuable for their ease and speed of calculation, but their assumptions break down for larger actuators and strains. It is preferable to combine the strengths of both these methods and fully characterize the actuators in the same manner that broader design spaces have been characterized for shape targeting without loading \cite{pikul2017stretchable, forte2022inverse}.

We take inspiration from research that has used Kevlar strain limiters to reinforce and shape the extension of silicone membranes \cite{sholl2021controlling, ceron2018fiber}. Unlike the membranes discussed above, the trajectory of these Fiber-Reinforced Elastomeric Enclosure (FREE) is entirely defined by the inextensible deformation of the fiber elements. The design space for force application with single-expansion FREE’s has been explored generally for slim cylindrical actuators with stiffer rubbers \cite{connolly2017automatic}, including in the presence of external loadings \cite{sedal2021comparison, bishop2015design}. These slender actuators can struggle under compressive loading due to buckling \cite{thomalla2022high}, while wider-based balloons have been shown to be useful for lifting \cite{sholl2021controlling, campbell2022electroadhesive, ambrose2023compact, devlin2020untethered} and are less susceptible to catastrophic buckling due to their  low slenderness and expanding cross-sectional area. Softer rubbers also allow the actuators in this work to operate at low pressures relative to the FREE community, with max pressures of 7.5~kPa.



Sequential experimentation and active learning provide a means of  collecting new data to minimize overall error of a machine learning model \cite{ren2021ALsurvey}. Such models are able to provide accurate predictions at very fast speeds at inference time (forward pass on order of ms). Learning has been used for inverse design of shape response for strain-limited membranes \cite{forte2022inverse}, but data-driven methods of characterization have underperformed \textit{energy} methods in the presence of external forces \cite{sedal2021comparison}. We aim to leverage active learning to explore the parameterized design space in a data-efficient way and reduce global model uncertainty. This will allow for the design of actuators targeted at specific lifting applications and reduced model error compared to theoretical, energy-based, methods.

In this work we characterize a parameterized class of actuator to design soft pneumatic actuators with optimized force application. To understand the response of the inflated system, we solve energy minimization relations that estimate the inflated shape of the actuator in the presence of a known external force. We then use active learning to collect an efficient dataset that spans the prescribed parameter space and that includes strains beyond the linear-elastic region of elastomer deformation. We train our neural network model to interpolate and predict force response between collected data and design parameters. We prescribe a target lift (height and force) trajectory for a single pressure sweep and obtain a membrane design as output. We demonstrate the utility of this model by using it to lift a mass along targeted trajectories as well as to  maximize lift height.


\section{Theoretical actuator modeling}
\label{sec: Theory}


\subsection{Elastomeric Thin Membrane}

\begin{figure} [htb]
    \centering
    \includegraphics[width= 0.7\linewidth]{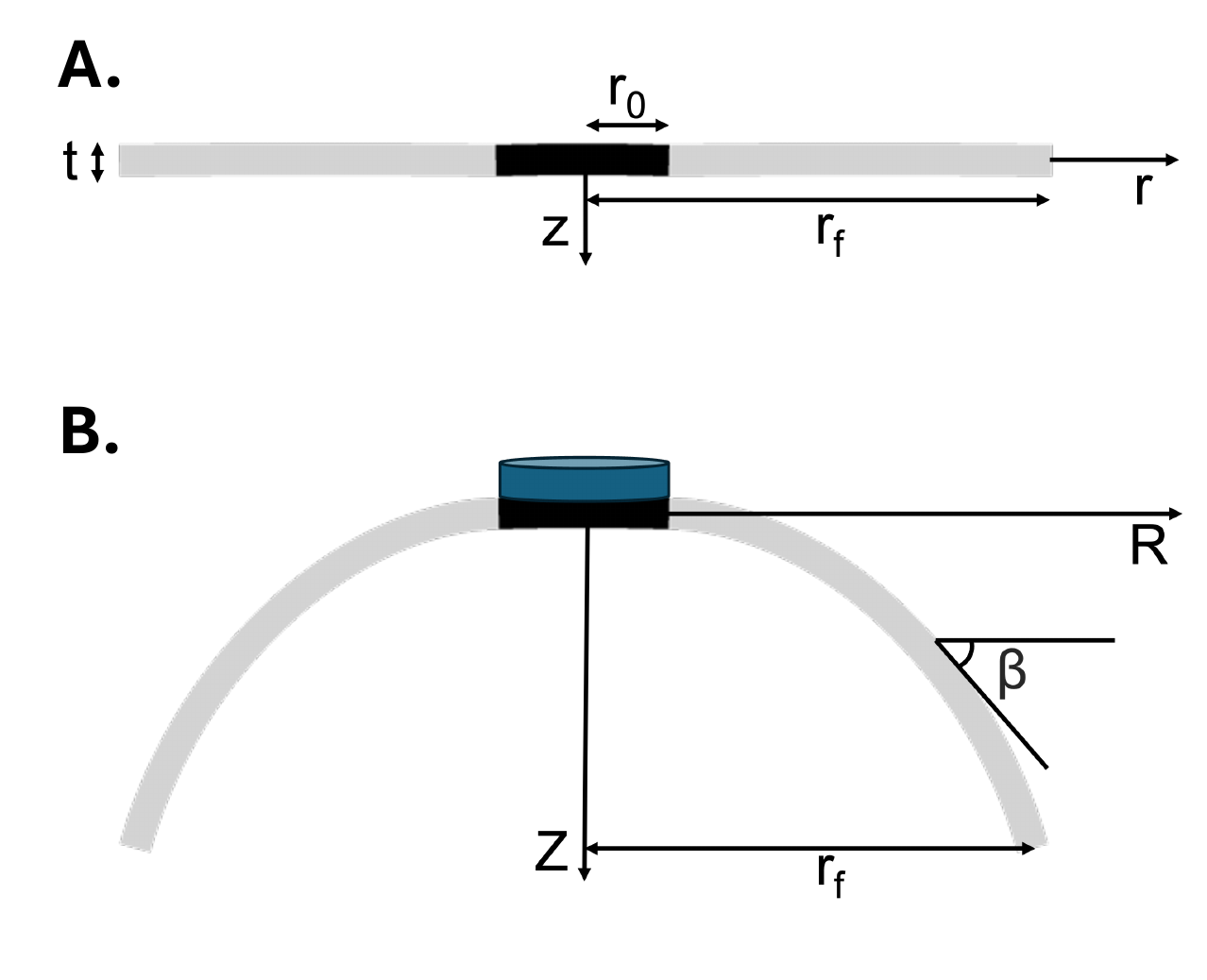}
    \caption{Schematic of the membrane (gray) with strain limiter (black) in the A. undeformed state, and B. deformed state while in contact with external force (blue).}
    \label{fig:Membrane schematic}
\end{figure}

We wish to determine the shape of a thin membrane upon inflation, with an external force applied  at membrane radius, $r$, ranging from $0\leq r \leq r_0$ as shown schematically in Fig.~\ref{fig:Membrane schematic}. 
Before deformation, the shape of the membrane that is not in contact with the force, using cylindrical coordinates, is given by $r_0\leq r \leq r_f$, and membrane height $z=0$. The region $0\leq r \leq r_0,$ which is in contact with the applied force, consists of low-strain material and we assume that there is no deformation in that region, only a translation of the region along the $z$ direction. We further assume that the membrane stays axisymmetric after the deformation and its shape is given by,
\begin{equation}
    \label{eq:membrane parametrization}
    R=R(r),\:\:\: Z=Z(r).
\end{equation}
From this, we can write the principal stretches, $\lambda$, as, 
\begin{equation}
    \label{eq:stretches}
    \lambda_1=\sqrt{R'(r)^2+Z'(r)^2},\:\:\:\lambda_2=\frac{R(r)}{r},\:\:\: \lambda_3=\frac{T}{t},
\end{equation}
where $t$ and $T$ are the thickness before and after deformation respectively, and primed variables denote a derivative with respect to $r.$ These principal stretches correspond to the directions of meridian, latitude, and normal to the deformed membrane respectively. We will further assume that the material is incompressible so that we can take $\lambda_3=1/(\lambda_1 \lambda_2).$ We can now write an energy for the system that assumes full contact between the contact plate and the unstretchable contact region as shown in Fig.~\ref{fig:Graphical_Abstract}B. 
\begin{equation}
\begin{aligned}
    \label{eq: energy}
    E=\int_{r_0}^{r_f}\left(2\pi\: r \:t \:W(\lambda_1,\lambda_2) - \pi\:p R^2\:Z' \right)dr~+ \\ \int_0^{r_0} \left( F Z'+2\pi\: r \:t \:W(\lambda_1,\lambda_2) - \pi\:p R^2\:Z' \right)dr,
\end{aligned}
\end{equation}

\noindent where $W$ is the strain energy per unit undeformed volume, $p$ is the gauge pressure inflating the membrane. In this study, we use the Gent model for the strain energy density function \cite{zhou2018evaluation}, which contains two constants, the shear modulus $\mu$ and the extension limit constant $J_m,$ which we find using uniaxial testing.
We set the first variation of the energy to zero and apply geometric relations based on $\beta(r),$ which we take to be the angle between the tangent to the membrane and the horizontal line at point $r,$ as shown in Fig. \ref{fig:Membrane schematic}B, \hbox{$R'(r)=\lambda_1(r) \text{cos}(\beta(r)),$} $\:\:\:Z'(r)=\lambda_1(r) \text{sin}(\beta(r))$. We write the equilibrium equations for the $r_0\leq r \leq r_f$ region as:
\begin{align}
&\frac{d\lambda_1}{dr}=\frac{W_2-\lambda_1 W_{12}}{r W_{11}}\text{cos}\beta+\frac{\lambda_2 W_{12}-W_1}{r W_{11}},\label{eq: equilibrium equations1}\\ &\frac{d\lambda_2}{dr}=\frac{\lambda_1 \text{cos}\beta-\lambda_2}{r},\label{eq: equilibrium equations2}\\
&\frac{d\beta}{dr}=\frac{\tilde{p}r \lambda_1 \lambda_2-W_2 \text{sin}\beta}{r W_1},\label{eq: equilibrium equations3}
\end{align}
where $\tilde{p}=p/t, \:W_1=\partial W/\partial \lambda_1, \:W_2=\:W/\partial \lambda_2,$ and \hbox{$W_{12}=\partial^2 W/\partial \lambda_1 \lambda_2.$}

The second term of the energy in Eq. \ref{eq: energy} only contributes to the boundary conditions at $r_0$ since there is no deformation to the shape in the $0\leq r < r_0$ region. The boundary conditions that come from the integration by parts give us a condition on the angle $\beta,$ at $r=r_0,$ so the initial boundary conditions needed to solve Eqns. \eqref{eq: equilibrium equations1}, \eqref{eq: equilibrium equations2}, and \eqref{eq: equilibrium equations3} are given by, 
\begin{align}
    \label{eq:initial conditions}
    &\lambda_1(r_0)=x,\\
    &\lambda_2(r_0)=1,\\
&\beta(r_0)=\text{ArcSin}\left(\frac{-F+\pi p\:r_0^2\lambda_2^2}{2\pi t\:r_0 W_1}\right)\Big|_{r=r_0}.
\end{align} 
The condition on $\lambda_2$ comes from the fact that the contact area and the plate are in full contact and there is no extension in that area. The condition on $\beta$ comes from the boundary conditions obtained during energy minimization. To solve the condition on $\lambda_1,$ we use the shooting method to find the value of $x$ that satisfies the condition $\lambda_2(r_f)=1.$ This final condition states that the membrane is fixed at the ends. Integrating the equilibrium equations from $r_0$ to $r_f$ obtains the shape of the deformed membrane. 

\subsection{Concentrically Strain Limited Thin Membrane}
\begin{figure*} [!t]
    \centering
    \includegraphics[width=\textwidth]{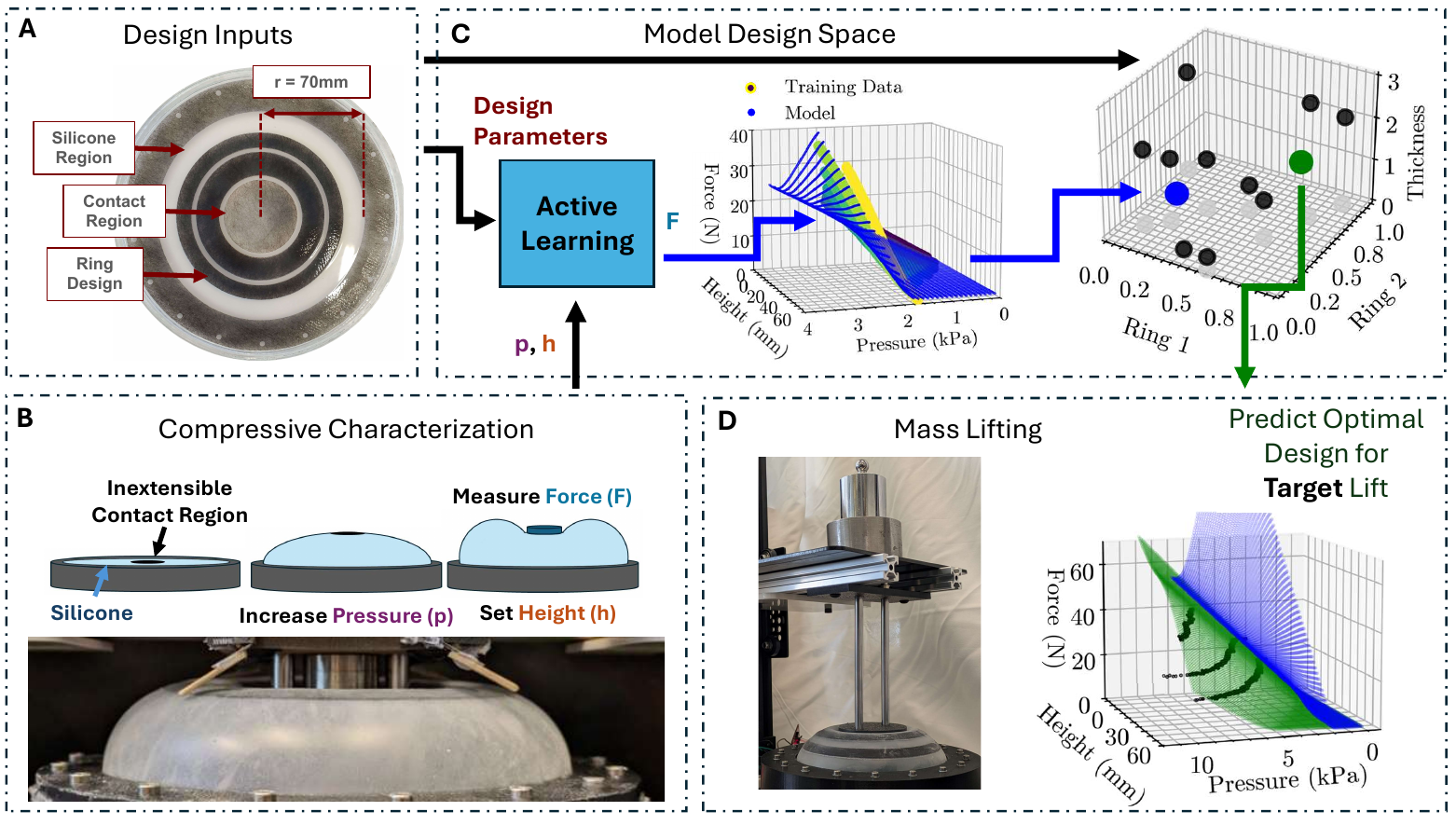}
    \caption{A. Design parameters for example membrane. B. Compressive Testing (top) Testing procedure for each membrane: expansion from a flat plane into a set height load-cell. Procedure is repeated for different heights. (bottom) Example of physical test measuring force for varying pressure at a set height. C. (left) Pressure, height, and six design parameters predict force output for a given membrane. (middle) Model (blue) overlaid with force-pressure training data at varying heights (0-70mm from purple to yellow). (right) 3-D visualization of 6-D design space. Example training set points in black, model from C in blue, model from D in green. D. Optimization for lifting task. (left) Physical testing to verify lift trajectories at a given mass. (right) Model planes for (blue) training parameters and (green) design parameters optimized to hit target trajectories (trajectories in black).}
    \label{fig:Graphical_Abstract}
\end{figure*}

To span our actuator design space, we want to find the deformed shape of a membrane made up of elastic material that also includes strain limiting rings (see Fig.~\ref{fig:Graphical_Abstract}A). To do this, the energy in Eq. \eqref{eq: energy}\: will involve an extra integral for each additional piece we add to the system, with the strain energy density function, $W,$ being different for the elastic material and the strain limiting rings. We have to solve Eqns. \eqref{eq: equilibrium equations1}, \eqref{eq: equilibrium equations2}, and \eqref{eq: equilibrium equations3} for each of the material pieces separately, with the initial boundary conditions at each piece being related through the boundary condition at the point of contact as well as the final boundary condition $\lambda_2(r_f)=1.$ Due to the stiffness of the differential equations for the heterogeneous material membranes, however, solving this final boundary condition is unsuccessful for 25\% of membranes. The successful subset of solutions are used and the results are discussed in Sec. \ref{subsec: ringed model performance}.
\section{Design space and experimental modeling}

\subsection{Design Space}

While soft pneumatic actuators can take a very large number of configurations, this work focuses on a class of soft pneumatic actuator defined as a thin (between 1 and 3mm), circular membrane of radius 70mm made from EcoFlex 00-30 rubber and reinforced with up to two axisymmetric rings (Soft n' Shear fabric). We ensure at least 10mm between the outer and inner radius of a ring and contact areas between 25.4 and 38.1mm. Our design space is an axisymmetric, finite, subset of all membrane-based SPA. 
An example membrane is seen in Figure \ref{fig:Graphical_Abstract}A-B. 


\subsection{Data Collection}
We fabricate membranes using gravity molding of Ecoflex 00-30. Lasercut strain limiting rings and contact regions (Soft n' Shear) are applied  to the uncured silicone after degassing and before curing. Membranes are mounted on a 3d-printed pressure chamber, which houses an air pressure sensor (MPRLS0025) and an ESP32 microcontroller to wirelessly transmit pressure data. Air pressure is supplied by a  \hbox{4.5 V} DC air pump (ZR370-02PM) and released by a 12V solenoid valve (Plum Garden). A piezoelectric load cell measures forces via an acrylic contact plate the same radius as the membrane's contact region and load is limited to move vertically  with linear ball bearings.

We perform automated testing for each membrane where the contact plate is positioned vertically by linear actuators (Homend) and verified by a time-of-flight sensor (VL53L0X) between trials. A single trial consists of the activation of the pump and the subsequent inflation of the actuator. Inflation continues for ringed membranes until the internal gauge pressure reaches 4.3~kPa. Then the pump is deactivated and the solenoid valve releases air until internal pressure reaches atmospheric. If the membrane comes into contact with anything except the contact plate, the trial is completed. Three trials are performed per contact plate height, with eight heights, 0-70~mm, per membrane. 
If the membrane does not burst, testing is repeated to a maximum pressure of 6.1~kPa. Ringless membranes were given no pressure limit and allowed to inflate until contact with the test rig.

Data from each trial includes time, force, pressure, time of flight height data (left height, right height), flow rate, and contact with test rig (binary), material type, nominal thickness, radius, contact radius, test year, test month, test day, curing rack (A/B), contact plate nominal height, trial number, thicknesses (from destructive testing of some membranes), and ring data. While deflation data is recorded in some cases, exclusively inflation data is used in model training and verification. 
The data is also sorted into a learning-friendly format as a dictionary with key values matching membrane design parameters. 
Dictionary data in the form of a .pkl file can be found in the Github repository (see Data Availability section).

\subsection{Model Architecture}

\begin{figure}[!htb]
    \centering
    \includegraphics[width=0.95\linewidth]{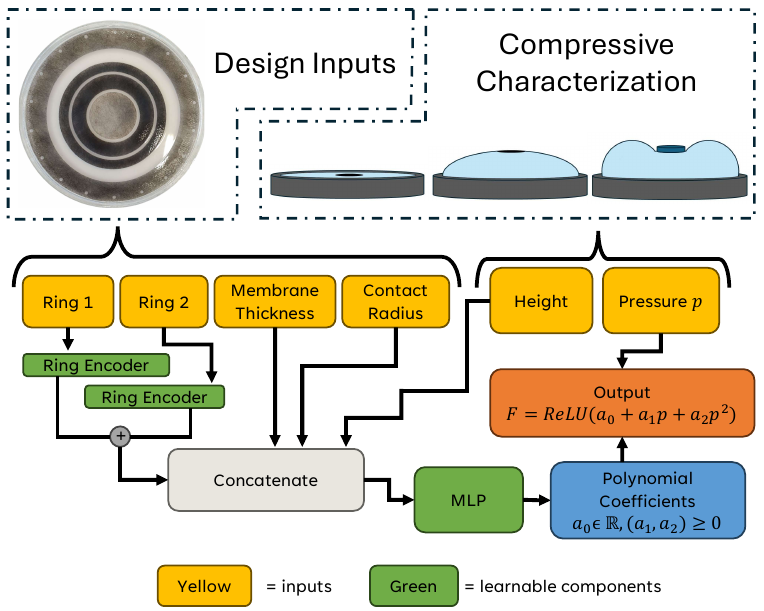}
    \caption{Model Architecture. (Top) Characterization data, pressure and height, and design inputs, ring radius and width, membrane thickness, and contact radius, are inputs to the actuator model. (Bottom) The  model solves for  force output 
    relative to pressure. 
    }
    \label{fig:model-architecture}
\end{figure}

Our model takes as an input a vector $M\in \mathbb{R}^6$ which encodes   six membrane design parameters: contact radius, membrane thickness, and four parameters describing ring locations (two for each ring, representing ring position and thickness), with NaN to indicate a lack of ring. Example formatting of the vector $M$ can be seen in Table \ref{tab: Mems}. The model also takes in object displacement (height) $h\in\mathbb{R}$ from the initial plane of the actuator and actuator internal gauge pressure $p\in\mathbb{R}$. In order to use the same model on ringed and ringless membrane designs, we employ a "Ring Encoder", which converts ring parameters (or their absence, indicated by NaN) into a latent representation. The two latent representations are then added together, which ensures that the model gives identical predictions no matter which ring is considered "Ring 1" and which is considered "Ring 2". The ring information is then concatenated to membrane thickness, contact radius, and height before being fed to a Multi-Layer Perceptron (MLP) neural network.

The final output of our model predicts the external force $F\in \mathbb{R}$ at distance $h$ from the membrane when pressure $p$ is being applied. Using this model, we are able to interpolate between collected pressure-height-force data for tested membranes (Figure \ref{fig:Graphical_Abstract}D) and between design parameters to predict force response of untested membrane designs.

Our model can be seen as a special case of the Operator Leaning framework \cite{li2020fno, lu2021deeponet}, where either the output or input of a neural network model is a function, potentially living in infinite-dimensional space. It takes as input the membrane design $M$ and a specific height $h$, and outputs a function $F_{M,h}:\mathbb{R}\to\mathbb{R}$ that computes the external force at a given pressure. That is $F_{M,h}(p)$ is the force applied by membrane $M$ at height $h$ when pressure $p$ is being applied. During our ablations we considered different forms of this function, including a learnable basis via the DeepONet strategy \cite{lu2021deeponet}, or explicitly imposing a linear, quadratic or cubic polynomial assumption (quadratic shown in Fig. \ref{fig:model-architecture}).

\subsection{Active Learning}


In order to carry out Active Learning (AL) \cite{ren2021ALsurvey}, we must first be able to quantify the epistemic (model) uncertainty of our predictions under the operator learning framework \cite{guilhoto2024neon}. We do this by using a Randomized Prior Network (RPN) ensemble of independent networks \cite{osband2018rpn, yang2022scalable}. Each of the $N\in\mathbb{N}$ members of the ensemble combines a prior and a trainable network to output a different prediction $F_{M,h}^1(p),F_{M,h}^2(p),\dots, F_{M,h}^N(p)$, which are averaged in order to compute the final prediction $F_{M,h}(p)=\frac{1}{N}\sum_{i=1}^NF_{M,h}^i(p)$. The epistemic uncertainty is computed as the standard deviation of these predictions, with a small value indicating agreement between members of the ensemble (low epistemic uncertainty), and a large value indicating disagreement (high epistemic uncertainty). The active learning procedure is then carried out by selecting the membrane $M\in\mathbb{R}^6$ for which predictions on average have the highest possible value of epistemic uncertainty. Testing such a membrane and training the model with this newly acquired data then decreases the uncertainty for this membrane and other designs similar to it, since the model is now trained with the experiments from this specific design. We carry out this procedure iteratively, until enough membranes are acquired.

In particular, we carry out active learning in the parallel setting, where at each iteration $q=2$ membranes are obtained simultaneously. We determine the best pair of membranes to collect by selecting the designs that present largest uncertainty in predictions at selected heights $h_1,\dots, h_{N_h}$ and pressures $p_1,\dots,p_{N_p}$. This is quantified via the acquisition function
\begin{equation}\label{eq: al-acquisition}
    \alpha(M_1, M_2) = \sum_{l=1}^N\max_{k\in\{1,2\}} \left[ \sqrt{\sum_i \sum_j \left(F_{M_k,h_i}(p_j)-F_{M_k,h_i}^l(p_j)\right)^2} \right]
\end{equation}
which is maximized using the L-BFGS optimizer with several starting points, then taking the best optimized value overall. Note that the expression in Eqn. (\ref{eq: al-acquisition}) is easily extendable to settings where we wish to collect an arbitrary number $q\in\mathbb{N}$ of membranes at each iteration, instead of only two.

\section{Results: Characterization \& Modeling}

\begin{figure} [!ht]
    \centering
    \includegraphics[width=\linewidth]{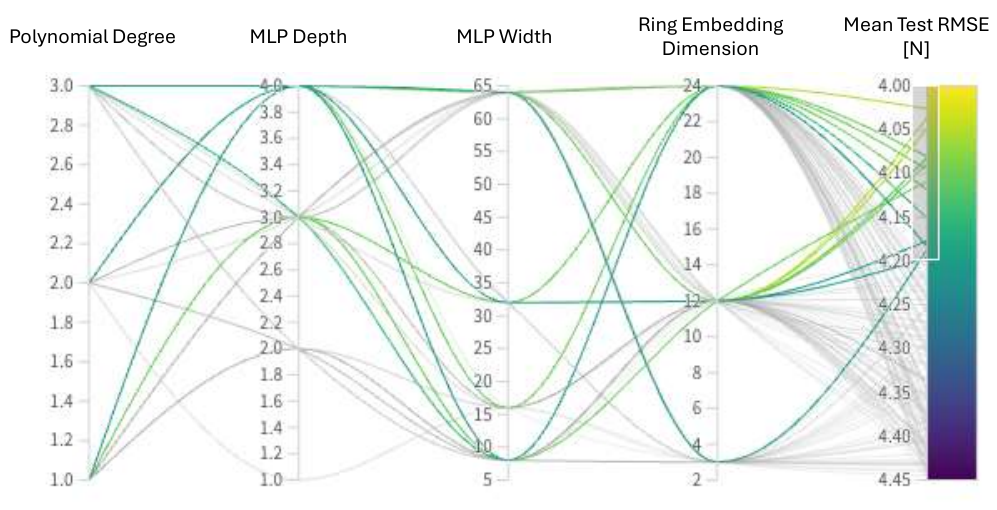}
    \caption{Model RMSE [N] of best performing model hyperparameters tracked to the corresponding parameters: output force polynomial degree relative to pressure, multi-layer perceptron depth and width, and ring encoder MLP width.}
    \label{fig: hyper-sweep}
\end{figure}

\subsection{Actuator Performance}
The membranes central to these actuators use shore hardness 00-30 silicone rubber. This combined with the 70~mm actuator radius allows them to operate at pressures under 7.5~kPa and apply forces reaching 103~N. Maximum force occurred with a ringless membrane of thickness 2~mm and contact radius 38.1~mm at a height of 50~mm. We characterized each actuator's displacement up to 70~mm when possible, and heights up to 79~mm were reached during mass lifting. Without taking into account the membrane properties, the maximum contact area tested, $45.6~cm^2$, would only be expected ($F=p*A$) to provide approximately 34~N of force. 

We fabricated 28 different membranes, of which 22 were ultimately used in model training. Other membrane data was discarded primarily due to stiffening element delamination or development of a hole in the membrane. These modes of failure occurred exclusively at boundaries between unstiffened and stiffened regions of silicone. Fracture due to expansion past the rubber's strain limitation did not disqualify a membrane, though the we removed the trial involving the fracture from that membrane's dataset. The resulting trimmed dataset included 188,318 individual data-points. Individual membrane designs each had between 1,907 and 16,837 data-points.


\subsection{Model Performance (Ringless)} \label{subsec: ringless model performance}
We compare the performance of the MLP-based pipeline to the theoretical (energy-minimization) model and a baseline that relies on the curve-fit function from Scikit-learn. The theoretical model, described in Section~\ref{sec: Theory}, requires empirical material constants to define the strain energy of the silicone. We performed and characterized uniaxial strain testing on rectangular samples of EcoFlex 00-30, similar to \cite{Marechal2020}, to solve for values of shear modulus $\mu = 31.5 kPa$ and strain-limiting factor $J_m = 39.6$. 


We perform a $k=3$ $k$-fold validation with pressure, height, and membrane design parameters as inputs and force as an output across the six ringless membranes. That is, for each of the $k=3$ folds, we pick two membranes as our test set and use the remaining 4 for training, cycling through the 3 different groups and taking the average test error across the 3 folds. The methods predicted force with the following RMSE: Neural Network (NN): 6.4~N, Simulation: 5.9~N, Curve-Fit Baseline: 5.6~N. With the addition of the ringed data, the NN RMSE dropped to 5.1~N. 
The maximum force seen in the lift trials was 103~N, of which these RMSE represent between 5.0\% (5.1~N) and 6.2\% (6.4~N).

\subsection{Model Performance (Ringed)} \label{subsec: ringed model performance}
We optimize model hyper parameters for the entire dataset by sweeping through: form of $F_{M,h}:\mathbb{R}\to\mathbb{R}$, width/depth of network, latent ring representation, and number of ensemble networks. We compare the Root-Mean-Squared-Error (RMSE) in model force relative to relevant test-set characterization data in a $k=11$ $k$-fold cross-validation across 22 membranes. The highest performing hyper parameter combinations are shown in Fig.~\ref{fig: hyper-sweep}.

The best-performing hyper-parameters result in a RMSE of 4.0~N (4\% of max seen force) across ringed and ringless membranes, and decrease RMSE by 5\% relative to the worst permutations. The model generally performs best with $F_{M,h}$ as polynomial degree 1, MLP depth of 3 or 4, and ring embedding model dimension of 12 or 24. The MLP width doesn't have a large impact on result. Neither does the existence or parameters of a separate MLP for ring pre-processing.

It is difficult to baseline these values, as both the energy method simulation and curve-fit approach used for ringless trials fail to effectively deal with the paramaterized ring values in all cases. The simulation was successful for 75\% of ringed membranes and the average RMSE across both ringed and ringless membranes was 4.4~N. The curve-fit needed to be optimized separately with different numbers of input variables for ringed and ringless membranes, and the average RMSE across both sets was 8.4~N.

\section{Design for Open-Loop Applications}


Once the NN model is trained, we use it to predict actuator performance in lifting tasks. We define a lift trajectory by the change in height of the object being lifted, the force applied by the actuator, and the pressure inflating the actuator. Our model therefore predicts the effect of design parameter choices on lift trajectories. Experimental lifts are performed with a 1
degree-of-freedom (DoF) test stage that is constrained by gantry plates. Masses are placed on the gantry and force is transmitted to the actuator via a contact plate (see Fig. \ref{fig:Graphical_Abstract}.D). 

We first verify model predictions on two membranes (one with and one without rings) from the training set. We model the expected lift trajectories for three target masses, 1.5kg, 2.5kg, and 4kg, and choose three points along each trajectory as target way-points. The points are chosen by the associated heights: 5, 40, and 50 mm. These heights were chosen because they represent areas of the trajectory that vary greatly between designs. We perform each lift and solve the error (weighted L2 norm in pressure and height) between the experimental trajectory and each way-point. Force is not included in this error metric because it is prescribed by the chosen mass. The total error between a membrane and the target points is defined as the root-mean square of the nine (3 mass x 3 way-point) errors:
\begin{equation} 
    \label{eq:RMSE}
    \footnotesize
    RMSE = \sqrt{\frac{1}{n}  \sum_i\left[ \left(\frac{(p_{exp}-p_{target(i)})}{p_{max}}\right)^2 + \left(\frac{(h_{exp}-h_{target(i)})}{h_{max}}\right)^2 \right]}    
\end{equation} 
\noindent where $p_{max}$ and $h_{max}$ are the limits of the space in which we model membrane forces: 10~kPa and 50~mm. $p_{exp}$ and $h_{exp}$ represent the respective pressure and height values from the experimental trajectory that minimize the individual error.

\begin{figure} [!htb]
    \centering
    \includegraphics[width= \linewidth]{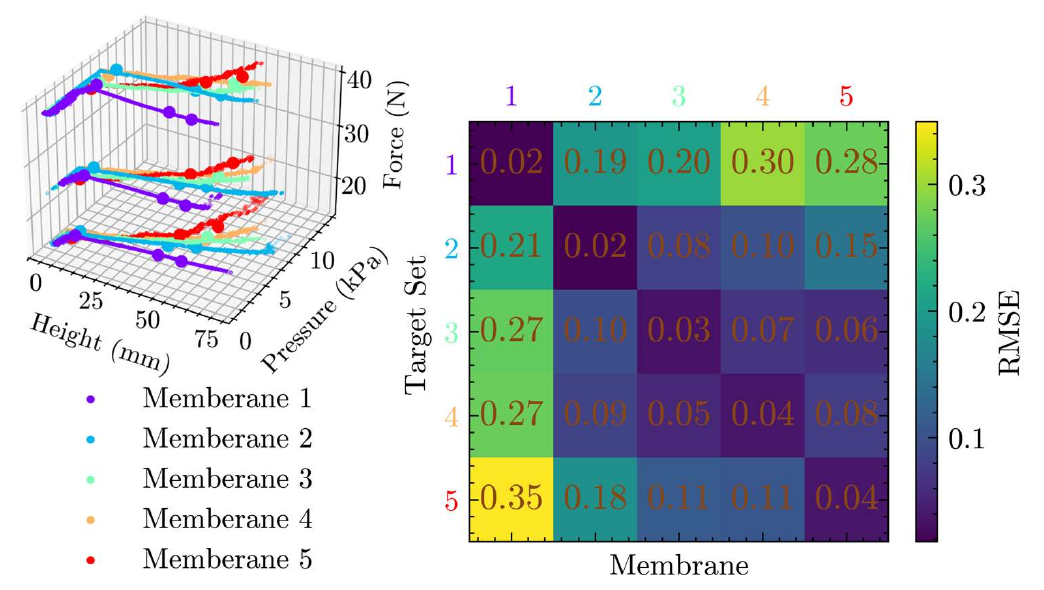}
    \caption{Experimental trajectories: (left) Pressure-height-force data from lifts at three different masses for each of five membranes (lines) with target points for each trajectory (circles). (right) RMSE between experimental trajectories and target way-points for each pairing of membrane and target.}
    \label{fig: Exp_Traj_Error}
\end{figure}

The scaled RMSE for the in-set ringless and ringed membrane trajectories are both 0.02. We then designate trajectories for three membrane designs not in the training set. We choose these membranes specifically to span the pressure-height-force state space. We designate nine way-points for each membrane based on the model. We perform mass testing and error calculation as designated above. The scaled RMSE for these membranes are 0.03 and 0.04. Parameters for all five membranes are listed above the dashed line in Table~\ref{tab: Mems}. Ring radius describes the radius at the center of the ring, ring width is the distance from that radius to the inner or outer radius ($r_{outer} - r_{inner} = 2 \cdot width$). For testing, the thickness values are restricted between 2.0 and 3.0~mm to minimize popping events. All other parameter ranges remain the same.
The experimental trajectories and their target way-points are shown in Fig.~\ref{fig: Exp_Traj_Error}. The associated errors are also shown in Fig.~\ref{fig: Exp_Traj_Error} - the main diagonal elements represents error along the trajectory for which each membrane was designed, off-diagonal elements represent each membrane's error related the the other membranes' trajectories. The decrease in error for a targeted membrane relative to the average of the other membranes' errors range from 69\% to 93\%.

\begin{center}
\begin{threeparttable}
\caption{Mass lift membrane design parameters [mm]}

\begin{tabular}{>{\centering\arraybackslash}p{1.25cm}|
                >{\centering\arraybackslash}p{0.95cm}|
                >{\centering\arraybackslash}p{0.95cm}|
                >{\centering\arraybackslash}p{0.95cm}|
                >{\centering\arraybackslash}p{0.95cm}|
                >{\centering\arraybackslash}p{0.95cm}}
\toprule
\makecell{ \\Thickness} & \makecell{Contact\\Radius} & \makecell{Ring\_1 \\ Radius} & \makecell{Ring\_1 \\ Width} & \makecell{Ring\_2 \\ Radius} & \makecell{Ring\_2 \\ Width} \\
\midrule
2.0  & 25.4 & nan   & nan  & nan   & nan  \\
2.0  & 25.4 & 49.0 & 5.0 & 62.0 & 5.0 \\
2.3  & 29.6 & 37.6 & 5.0 & 62.0 & 5.0 \\
2.0  & 28.0 & 45.6 & 5.0 & 60.3 & 6.7 \\
2.0  & 38.1 & 47.6 & 6.4 & 62.0 & 5.0 \\
\hdashline  
2.0  & 25.4 & 33.4 & 5.0 & 46.4 & 5.0 \\
2.0  & 31.9 & 46.0 & 5.0 & 59.0 & 5.0 \\
\bottomrule

\end{tabular}
\label{tab: Mems}
\end{threeparttable}
\end{center}

\vspace{6pt}

Using the inherent gradients and fast solve-time of the trained model, we are able to optimize for specific lift goals. As an example, we maximize lift height at given targets of pressure and force. We define the following posterior function~$\Pi$: $\Pi = -k_{force}*F_{error} - k_{pressure}*p_{error} + k_{height}*h_{min}$

\noindent where $k$ are scaling factors and $h_{min}$ is the 'smooth min' score (via the LogSumExp operation of three height values). Local minima are found using gradient descent and compared across 2,500 random starting points. We choose two sets of target points,  with forces across the three target masses (14.7 to 39.2~N). Target pressures are set at 6.9~kPa (set A) and 8.3~kPa (set B). The optimized parameters are listed below the dashed line in Table \ref{tab: Mems}.


We fabricate membranes based on the results of the optimization and test them as described for the five membranes above. We search the lift trajectories for points closest to the target force and pressure then record lift heights for each membrane at these points. A score is given based on the three heights, with a higher score matching a larger height. For each set of target points, the newest membranes had the highest score. Optimal membrane A reached a score of 29.4. Among the five prior membranes, four reached all target points for set A and they averaged a score of 14.4. Optimal membrane B reached a height score of 40.8 for target set B. Two of the five prior membranes were able to reach the set B target pressure-force combinations, these two averaged a score of 25.7.


\section{Discussion}
This paper uses a machine learning model to inform our design decisions within a parameterized design space for a force application task. We model the theoretical mechanisms governing the expansion of silicone-based SPA and develop a custom Neural Network (NN) architecture for exploring and quantifying the effects of the parameterized design space. The outputs of this model were verified and compared against two other types of models using a dataset gathered with automated experimentation. The trained model is used to define trajectory waypoints for membranes both in and out of the training set, confirmed with experimental mass lifts, and then to optimize for a specific lift output (maximize height). 


The SPA we characterize in this study provide performance that could enable affordable means of force application for human motion. An actuator with the footprint of an adult head and a contact area smaller than an adult fist, backed by a 5~V battery and a portable air pump, is shown to provide forces over 100~N over a workspace of over 50~mm. Using just 22 membrane characterizations, we are able to train a model that provides 4.0~N RMSE across the entire parameterized design space. For a given force, we can design an actuator that will reach multiple points along a chosen pressure-height trajectory within approximately 4\% error in height and pressure. These metrics indicate potential for performing pre-meditated lifting trajectories via a simple pressure sweep.


We can predict the way in which embedding concentric stiffening elements will alter the lift trajectory of a silicone membrane, proven by the small predicted error described in Section \ref{subsec: ringed model performance}. It is important to contextualize the performance of our NN model with respect to state of the art theoretical models and off-the-shelf data-regression algorithms. We note that the simpler regression methods are not able to handle combined ringed and ringless data in the way that our NN model does. This and their large errors for ringed cases may indicate that they aren't a good choice for complex design parametrization. Similarly, the rigidity of the boundary condition on the energy minimization approach prevents its success in some ringed cases. Therefore for the ringed design space chosen for this work, the complexity of a NN appears to be warranted. Furthermore, once the NN model is trained, it is just as quick, if not quicker, to query as the other options.

When optimizing for lift height, parameters were often shifted to the edges of our design space. Specifically, the model recommended designs where the rings were as close as possible to either the contact area or the outer housing. Our manufacturing techniques don't allow for  high precision in the placement of the rings, but automated methods might allow us to relax design parameter constraints. This would allow us to answer, for instance, what distance between rings maximizes lift height at different loadings.

Within the raw data we see the stiffening elements do not increase maximum force output or workspace of the actuators within the tested pressure range, and that they increase the relative pressure needed for a given force or height output. Adding more stiffening elements also increases the potential for membrane fracture due to increased boundary areas. Though neither was studied extensively in this work,  this may decrease both the safe operating pressure for the membranes and their cycle life. While far more lift trajectories with ringed membranes are possible, a system designed using only ringless membranes may survive rougher usage.

Some of the most interesting performance within the collected dataset occur within unconstrained membranes. These membranes converge in pressure and force across different heights. This convergence is due to the hyperelastic constitutive relations of the silicone, and could be an important characteristic in lifting with this class of actuator. This convergence point also represents a difficulty for theoretical simulations, and comparative error was much higher at forces above this convergence point.

While the results of this particular membrane class are encouraging, the ability to effectively model any parameterized actuator class with a relatively small (n=22) dataset has wider implications within SPA and soft robotics generally. SPA researchers are using a variety of of multi-material designs to enable different interactions with the world. Active learning could help speed the exploration of these design spaces and lead to precisely tailored designs. Offloading control complexity from the pneumatic input to the mechanical design can also enable us to move toward inexpensive and untethered robots with useful force and motion outputs. 

\section{Conclusions \& Future Work}

We collect an active learning enabled n=22 dataset comparing force, height, and pressure across the design space for 70~mm radius concentric ring strain limited silicone soft pneumatic actuators. Actuators in this design space are found to apply forces over 100~N and reach heights of 80~mm. This dataset allows for an empirical model with less error than theoretical energy methods and naive curve-fit models. The empirical model is fully differentiable, which we leverage for design optimization in a height-maximization mass lifting task.

While there is potential for these actuators to be useful individually given their relatively high force and displacement outputs, we foresee a larger potential impact from their use in parallel. Because this model allows us to design for a given force/height pairing at a given pressure, we can connect multiple actuators to a single pressure source and use their embodied response to prescribe their lift trajectories. If we can model the, for instance, rigid body they are lifting, we can also model the kinematics of the lift from a single pressure source. While membranes displayed low hysteresis in our characterization testing, additional work will be required to see what additional force-height-pressure planes can be reached once a set of actuators is attached in parallel.

This model also allows for us to estimate the effect of (de)activating specific strain limiters, so long as the initial and final strain states remain in the design space. This, combined with variable limiters like those used in \cite{campbell2022electroadhesive} could be used to alter lift trajectories in real-time for open- or closed-loop lifting. 

While we consider the characterization data modeled in this study to be the most pertinent to soft pneumatic actuator lifting, there were additional datastreams recorded that will be shared on Dryad, including volumetric flow and video of membrane expansion. We hope this data can be useful for better understanding the properties of inflated membranes undergoing concentric one-dimensional loading or as a comparison for improving analytical analyses.

A possible future direction of research using this data is to develop a multi-fidelity model that leverages both experimental and simulated data. Such a model could use large amounts of high-throughput simulated data and calibrate predictions based on a small number of real-world experiments. Such a model has the potential to drastically reduce uncertainty and increase overall predictive accuracy.

\section*{ACKNOWLEDGMENT}

Thanks to Jason Matthew, R. Daelan Roosa and \hbox{Dr. Kevin Turner} for help in testing.  This work was supported in part by NSF EFRI award \#1935294 and NSF NRT award \#2152205. 

\section*{DATA AVAILABILITY}

The majority of the work above was coded in Python, including in Jupyter Notebooks. This code, and a .pkl file containing a dictionary of all the test data used herein, is available on Github: \url{https://github.com/gmcampbell/SPA\_Design}. 

The complete dataset, including tests not included in this paper and video of testing when applicable, is freely available on Dryad: \url{https://doi.org/10.5061/dryad.jsxksn0mt}.
\footnote{During review, this Dryad dataset is available only at: \url{http://datadryad.org/share/TuwV6VbkcyTUOiPFrrhi9ycvm_MKxX_Pn0WqU3sKE9A}}






\bibliographystyle{IEEEtran}
\bibliography{bibliography}

\end{document}